\documentclass[runningheads,a4paper]{llncs}

\usepackage{amssymb}
\usepackage{graphicx}
\usepackage{times}

\usepackage{comment} 
\usepackage{graphicx} 
\usepackage{epstopdf} 
\usepackage{amsmath} 
\usepackage{mathtools} 
\usepackage{mathrsfs} 
\usepackage{scrextend} 
\usepackage{url} 
\usepackage{epstopdf}
\usepackage{bm} 
\usepackage{amssymb} 
\usepackage{color}
\usepackage{flushend} 

\makeatletter
\def\hlinew#1{%
	\noalign{\ifnum0=`}\fi\hrule \@height #1 \futurelet
	\reserved@a\@xhline}
\makeatother

\newcommand{\tabincell}[2]{\begin{tabular}{@{}#1@{}}#2\end{tabular}} 
\usepackage{url}
\newcommand{\keywords}[1]{\par\addvspace\baselineskip
\noindent\keywordname\enspace\ignorespaces#1}

\begin{document}

\mainmatter  

\title{Deep Multi-instance Networks with Sparse Label Assignment for Whole Mammogram Classification}

\titlerunning{Deep MIL with Sparse Label Assignment for Whole Mamm Class.}

%
%
\author{Wentao Zhu, Qi Lou, Yeeleng Scott Vang, and Xiaohui Xie} 
%
\authorrunning{W. Zhu et al.}
\institute{Dept. of Computer Science, University of California, Irvine \\ \{wentaoz1, xhx\}@ics.uci.edu, \{qlou, ysvang\}@uci.edu}
%
%
\toctitle{Lecture Notes in Computer Science}
\tocauthor{Authors' Instructions}
\maketitle
\begin{abstract}
  Mammogram classification is directly related to computer-aided diagnosis of breast cancer. Traditional methods rely on regions of interest (ROIs) which require great efforts to annotate. Inspired by the success of using deep convolutional features for natural image analysis and multi-instance learning (MIL) for labeling a set of instances/patches, we propose end-to-end trained deep multi-instance networks for mass classification based on whole mammogram without the aforementioned ROIs. We explore three different schemes to construct deep multi-instance networks for whole mammogram classification. Experimental results on the INbreast dataset demonstrate the robustness of proposed networks compared to previous work using segmentation and detection annotations. \footnote[1]{The code can be downloaded from https://github.com/wentaozhu/deep-mil-for-whole-mammogram-classification.git.}
\keywords{Deep multi-instance learning, whole mammogram classification, max pooling-based MIL, label assignment-based MIL, sparse MIL} 
\end{abstract}
\section{Introduction}\label{sec:intro}
According to the American Cancer Society, breast cancer is the most frequently diagnosed solid cancer and the second leading cause of cancer death among U.S. women~\cite{acs}. Mammogram screening has been demonstrated to be an effective way for early detection and diagnosis, which can significantly decrease breast cancer mortality~\cite{oeffinger2015breast}. Traditional mammogram classification requires extra annotations such as bounding box for detection or mask ground truth for segmentation~\cite{varela2006use,carneiro2015unregistered,jiao2016deep}. Other work have employed different deep networks to detect ROIs and obtain mass boundaries in different stages \cite{dhungel2016automated}. However, these methods require hand-crafted features to complement the system~\cite{kooi2017large}, and training data to be annotated with bounding boxes and segmentation ground truths which require expert domain knowledge and costly effort to obtain. In addition, multi-stage training cannot fully explore the power of deep networks. 

Due to the high cost of annotation, we intend to perform classification based on a raw whole mammogram. Each patch of a mammogram can be treated as an instance and a whole mammogram is treated as a bag of instances. The whole mammogram classification problem can then be thought of as a standard MIL problem. Due to the great representation power of deep features~\cite{greenspan2016guest,zhu2016adversarial,zhu2016co,zhu2015hierarchical}, combining MIL with deep neural networks is an emerging topic. Yan et al. used a deep MIL to find discriminative patches for body part recognition~\cite{yan2016multi}. Patch based CNN added a new layer after the last layer of deep MIL to learn the fusion model for multi-instance predictions~\cite{hou2015patch}. Shen et al. employed two stage training to learn the deep multi-instance networks for pre-detected lung nodule classification~\cite{shen2016learning}. The above approaches used max pooling to model the general multi-instance assumption which only considers the patch of max probability. In this paper, more effective task-related deep multi-instance models with end-to-end training are explored for whole mammogram classification. We investigate three different schemes, i.e., max pooling, label assignment, and sparsity, to perform deep MIL for the whole mammogram classification task.
\begin{figure}[t]
	\setlength{\abovecaptionskip}{0.cm}
	\setlength{\belowcaptionskip}{-0.cm}
	\begin{center}
		\begin{minipage}{0.8\linewidth}
			\centerline{\includegraphics[width=\textwidth]{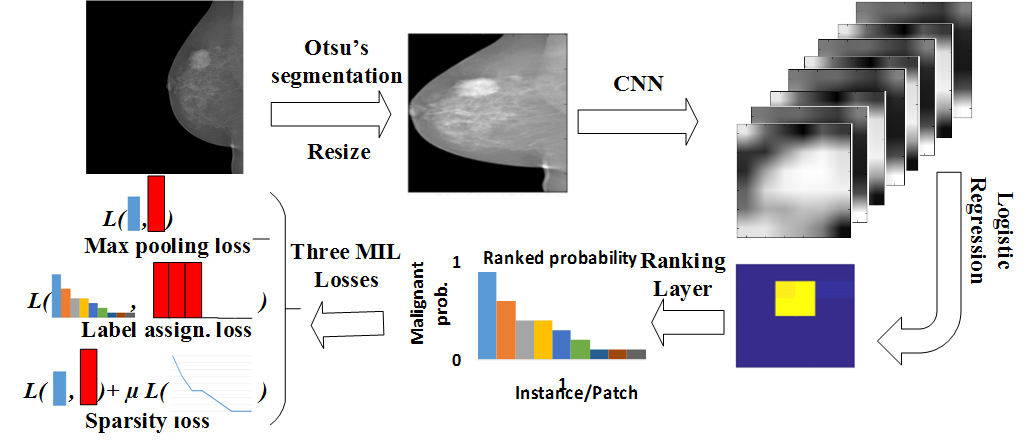}}
		\end{minipage}
		\caption{The framework of whole mammogram classification. First, we use Otsu's segmentation to remove the background and resize the mammogram to $227\times227$. Second, the deep MIL accepts the resized mammogram as input to the convolutional layers. Here we use the convolutional layers in AlexNet~\cite{krizhevsky2012imagenet}. Third, logistic regression with weight sharing over different patches is employed for the malignant probability of each position from the convolutional neural network (CNN) feature maps of high channel dimensions. Then the responses of the instances/patches are ranked. Lastly, the learning loss is calculated using max pooling loss, label assignment, or sparsity loss for the three different schemes.}
		\label{fig:framework}
	\end{center}
\end{figure}
\begin{figure}[t]
	\setlength{\abovecaptionskip}{0.cm}
	\setlength{\belowcaptionskip}{-0.cm}
	\begin{center}
		\begin{minipage}{0.245\linewidth}
			\centerline{\includegraphics[width=0.9\linewidth]{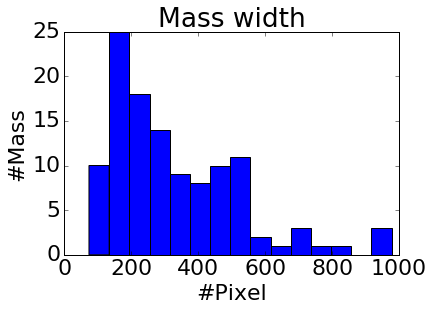}}
			\center{(a)}
		\end{minipage}
		\begin{minipage}{0.245\linewidth}
			\centerline{\includegraphics[width=0.9\linewidth]{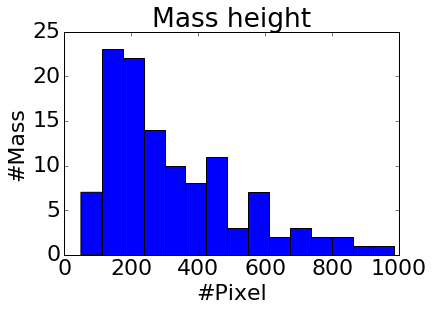}}
			\center{(b)}
		\end{minipage}
		\begin{minipage}{0.245\linewidth}
			\centerline{\includegraphics[width=0.9\linewidth]{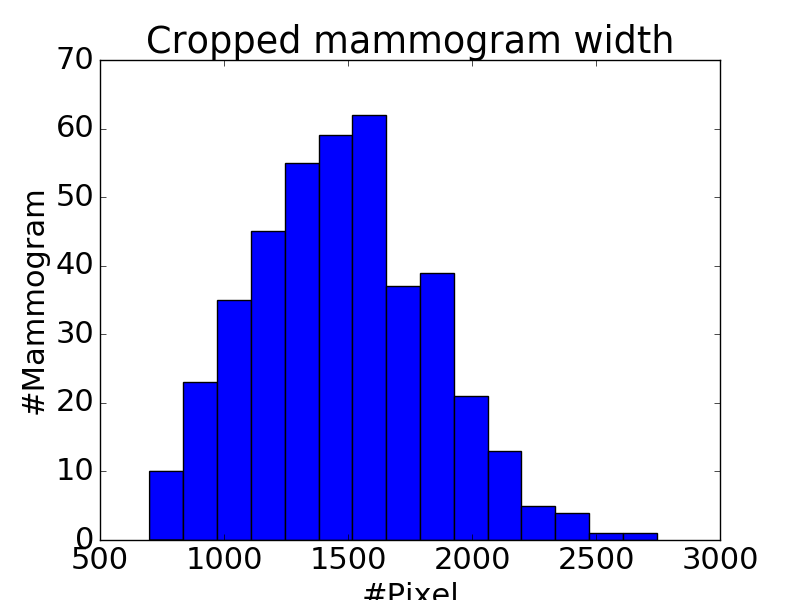}}
			\center{(c)}
		\end{minipage}
		\begin{minipage}{0.245\linewidth}
			\centerline{\includegraphics[width=0.9\linewidth]{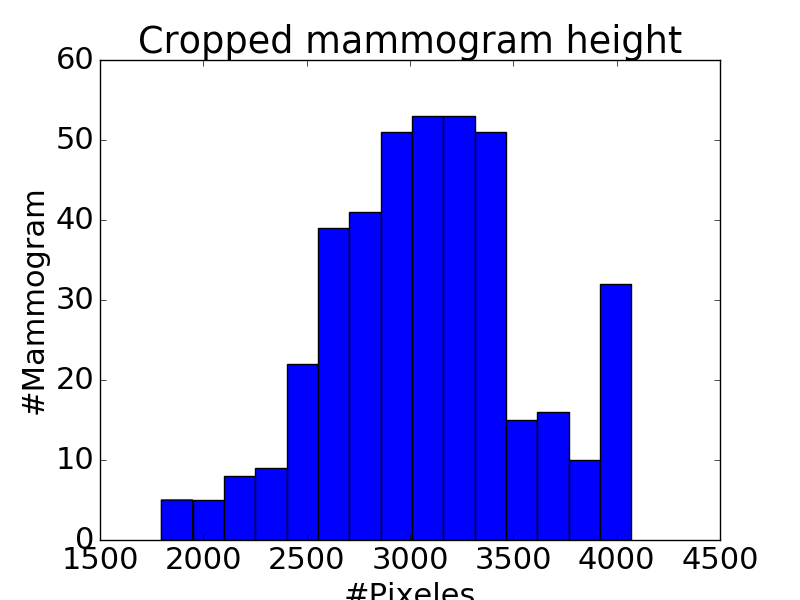}}
			\center{(d)}
		\end{minipage}
		\caption{Histograms of mass width (a) and height (b), mammogram width (c) and height (d). Compared to the size of whole mammogram ($1,474 \times 3,086$ on average after cropping), the mass of average size ($329 \times 325$) is tiny, and takes about 2\% of a whole mammogram. }
		\label{fig:mass}
	\end{center}
\end{figure}

The framework for our proposed end-to-end trained deep MIL for mammogram classification is shown in Fig.~\ref{fig:framework}. To fully explore the power of deep MIL, we convert the traditional MIL assumption into a label assignment problem. As a mass typically composes only 2\% of a whole mammogram (see Fig.~\ref{fig:mass}), we further propose sparse deep MIL. The proposed deep multi-instance networks are shown to provide robust performance for whole mammogram classification on the INbreast dataset~\cite{moreira2012inbreast}.
\section{Deep MIL for Whole Mammogram Mass Classification}\label{sec:dmlmamm}
Unlike other deep multi-instance networks~\cite{yan2016multi,hou2015patch}, we use a CNN to efficiently obtain features of all patches (instances) at the same time. Given an image $\bm{I}$, we obtain a much smaller feature map $\bm{F}$ of multi-channels $N_c$ after multiple convolutional layers and max pooling layers. The $(\bm{F})_{i,j,:}$ represents deep CNN features for a patch $\bm{Q}_{i,j}$ in $\bm{I}$, where $i,j$ represents the pixel row and column index respectively.

The goal of our work is to predict whether a whole mammogram contains a malignant mass (BI-RADS $\in \{4, 5, 6\}$ as positive) or not, which is a standard binary classification problem. We add a logistic regression with weights shared across all the pixel positions following $\bm{F}$, and an element-wise sigmoid activation function is applied to the output. To clarify it, the malignant probability of feature space's pixel $(i,j)$ is
\begin{equation}
\label{equ:cnn}
r_{i,j} = \text{sigmoid}(\bm{a} \cdot \bm{F}_{i,j,:} + b),
\end{equation}
where $\bm{a}$ is the weights in logistic regression, $b$ is the bias, and $\cdot$ is the inner product of the two vectors $\bm{a}$ and $\bm{F}_{i,j,:}$. The $\bm{a}$ and $b$ are shared for different pixel positions $i,j$. We can combine $r_{i,j}$ into a matrix $\bm{r} = (r_{i,j})$ of range $[0, 1]$ denoting the probabilities of patches being malignant masses. The $\bm{r}$ can be flattened into a one-dimensional vector as $\bm{r} = (r_{1}, r_{2}, ..., r_{m})$ corresponding to flattened patches $(\bm{Q}_{1}, \bm{Q}_{2}, ..., \bm{Q}_{m})$, where $m$ is the number of patches.
\subsection{Max Pooling-based Multi-instance Learning}\label{sec:maxpool}
The general multi-instance assumption is that if there exists an instance that is positive, the bag is positive~\cite{dietterich1997solving}. The bag is negative if and only if all instances are negative. For whole mammogram classification, the equivalent scenario is that if there exists a malignant mass, the mammogram $\bm{I}$ should be classified as positive. Likewise, negative mammogram $\bm{I}$ should not have any malignant masses. If we treat each patch $\bm{Q}_{i}$ of $\bm{I}$ as an instance, the whole mammogram classification is a standard multi-instance task.

For negative mammograms, we expect all the $r_i$ to be close to 0. For positive mammograms, at least one $r_i$ should be close to 1. Thus, it is natural to use the maximum component of $\bm{r}$ as the malignant probability of the mammogram $\bm{I}$
\begin{equation}
\label{equ:max}
p(y=1|\bm{I}, \bm{\theta}) = \max\{ r_1, r_2, ..., r_m\},
\end{equation}
where $\bm{\theta}$ is the weights in deep networks.

If we sort $\bm{r}$ first in descending order as illustrated in Fig.~\ref{fig:framework}, the malignant probability of the whole mammogram $\bm{I}$ is the first element of ranked $\bm{r}$ as
\begin{equation}
\label{equ:sortmax}
\begin{aligned}
&\{{r^\prime}_1, {r^\prime}_2, ..., {r^\prime}_m\} = \text{sort} (\{ r_1, r_2, ..., r_m\}), \\
&p(y=1|\bm{I}, \bm{\theta}) = {r^\prime}_1, \quad\text{and}\quad p(y=0|\bm{I}, \bm{\theta}) = 1-{r^\prime}_1,
\end{aligned}
\end{equation}
where $\bm{r}^\prime = ({r^\prime}_1, {r^\prime}_2, ..., {r^\prime}_m)$ is descending ranked $\bm{r}$. The cross entropy-based cost function can be defined as  
\begin{equation}
\label{equ:maxloss}
\mathcal{L}_{maxpooling} = -\frac{1}{N}\sum_{n=1}^{N} \log(p(y_n | \bm{I}_n, \bm{\theta})) + \frac{\lambda}{2} \|\bm{\theta}\|^2
\end{equation}
where $N$ is the total number of mammograms, $y_n \in \{0,1\}$ is the true label of malignancy for mammogram $\bm{I}_n$, and $\lambda$ is the regularizer that controls model complexity.

One disadvantage of max pooling-based MIL is that it only considers the patch ${\bm{Q}^\prime}_1$, and does not exploit information from other patches. A more powerful framework should add task-related priori, such as sparsity of mass in whole mammogram, into the general multi-instance assumption and explore more patches for training. 
\subsection{Label Assignment-based Multi-instance Learning}\label{sec:labelassign}
For the conventional classification tasks, we assign a label to each data point. In the MIL scheme, if we consider each instance (patch) $\bm{Q}_i$ as a data point for classification, we can convert the multi-instance learning problem into a label assignment problem.

After we rank the malignant probabilities $\bm{r} = (r_{1}, r_{2}, ..., r_{m})$ for all the instances (patches) in a whole mammogram $\bm{I}$ using the first equation in Eq.~\ref{equ:sortmax}, the first few ${r^\prime}_i$ should be consistent with the label of whole mammogram as previously mentioned, while the remaining patches (instances) should be negative. Instead of adopting the general MIL assumption that only considers the ${\bm{Q}^\prime}_1$ (patch of malignant probability ${r^\prime}_1$), we assume that 1) patches of the first $k$ largest malignant probabilities $\{{r^\prime}_1, {r^\prime}_2, ..., {r^\prime}_k\}$ should be assigned with the same class label as that of whole mammogram, and 2) the rest patches should be labeled as negative in the label assignment-based MIL.

After the ranking/sorting layer using the first equation in Eq.~\ref{equ:sortmax}, we can obtain the malignant probability for each patch
\begin{equation}
\label{equ:ppatch}
\begin{aligned}
p(y=1 | {\bm{Q}^\prime}_i, \bm{\theta}) = {r^\prime}_i, \quad\text{and}\quad p(y=0 | {\bm{Q}^\prime}_i, \bm{\theta}) = 1-{r^\prime}_i.
\end{aligned}
\end{equation}

The cross entropy loss function of the label assignment-based MIL can be defined
\begin{equation}
\label{equ:weightedlabelloss}
\begin{aligned}
\mathcal{L}_{labelassign.} = &-\frac{1}{mN}\sum_{n=1}^{N}  \bigg ( \sum_{j=1}^{k} {\log(p(y_n | {\bm{Q}^\prime}_j, \bm{\theta}))}+ \\&\sum_{j=k+1}^{m} {\log(p(y=0 | {\bm{Q}^\prime}_j, \bm{\theta}))}\bigg )+\frac{\lambda}{2} \|\bm{\theta}\|^2.
\end{aligned}
\end{equation}

One advantage of the label assignment-based MIL is that it explores all the patches to train the model. Essentially it acts a kind of data augmentation which is an effective technique to train deep networks when the training data is scarce. From the sparsity perspective, the optimization problem of label assignment-based MIL is exactly a $k$-sparse problem for the positive data points, where we expect $\{{r^\prime}_1, {r^\prime}_2, ..., {r^\prime}_k\}$ being 1 and $\{{r^\prime}_{k+1}, {r^\prime}_{k+2}, ..., {r^\prime}_m\}$ being 0. The disadvantage of label assignment-based MIL is that it is hard to estimate the hyper-parameter $k$. Thus, a relaxed assumption for the MIL or an adaptive way to estimate the hyper-parameter $k$ is preferred. 
\subsection{Sparse Multi-instance Learning}\label{sec:sparse}
From the mass distribution, the mass typically comprises about 2\% of the whole mammogram on average (Fig.~\ref{fig:mass}), which means the mass region is quite sparse in the whole mammogram. It is straightforward to convert the mass sparsity to the malignant mass sparsity, which implies that $\{{r^\prime}_1, {r^\prime}_2, ..., {r^\prime}_m\}$ is sparse in the whole mammogram classification problem. The sparsity constraint means we expect the malignant probability of part patches ${r^ \prime}_i$ being 0 or close to 0, which is equivalent to the second assumption in the label assignment-based MIL. Analogously, we expect ${r^\prime}_1$ to be indicative of the true label of mammogram $\bm{I}$.

After the above discussion, the loss function of sparse MIL problem can be defined
\begin{equation}
\label{equ:weightedsparseloss}
\mathcal{L}_{sparse} = \frac{1}{N}\sum_{n=1}^{N} \big ( -\log(p(y_n | \bm{I}_n, \bm{\theta})) + \mu \|\bm{r}^{\prime}_n\|_1 \big ) +\frac{\lambda}{2} \|\bm{\theta}\|^2,
\end{equation}
where $p(y_n | \bm{I}_n, \bm{\theta})$ can be calculated in Eq.~\ref{equ:sortmax}, $\bm{r}_n = ({r^\prime}_1, {r^\prime}_2, ..., {r^\prime}_m)$ for mammogram $\bm{I}_n$, $\|\cdot\|_1$ denotes the $\mathcal{L}_1$ norm, $\mu$ is the sparsity factor, which is a trade-off between the sparsity assumption and the importance of patch ${\bm{Q}^\prime}_1$.

From the discussion of label assignment-based MIL, this learning is a kind of exact $k$-sparse problem which can be converted to $\mathcal{L}_1$ constrain. One advantage of sparse MIL over label assignment-based MIL is that it does not require assign label for each patch which is hard to do for patches where probabilities are not too large or small. The sparse MIL considers the overall statistical property of $\bm{r}$

Another advantage of sparse MIL is that, it has different weights for general MIL assumption (the first part loss) and label distribution within mammogram (the second part loss), which can be considered as a trade-off between max pooling-based MIL (slack assumption) and label assignment-based MIL (hard assumption).
\section{Experiments}\label{sec:exp}
We validate the proposed models on the most frequently used mammographic mass classification dataset, INbreast dataset~\cite{moreira2012inbreast}, as the mammograms in other datasets, such as DDSM dataset~\cite{bowyer1996digital}, are of low quality. The INbreast dataset contains 410 mammograms of which 100 containing malignant masses. These 100 mammograms with malignant masses are defined as positive. For fair comparison, we also use 5-fold cross validation to evaluate model performance as~\cite{dhungel2016automated}. For each testing fold, we use three folds for training, and one fold for validation to tune hyper-parameters. The performance is reported as the average of five testing results obtained from cross-validation.

We employ techniques to augment our data. For each training epoch, we randomly flip the mammograms horizontally, shift within 0.1 proportion of mammograms horizontally and vertically, rotate within 45 degree, and set $50 \times 50$ square box as 0. In experiments, the data augmentation is essential for us to train the deep networks.

For the CNN network structure, we use AlexNet and remove the fully connected layers~\cite{krizhevsky2012imagenet}. Through CNN, the mammogram of size $227 \times 227$ becomes 256 $6 \times 6$ feature maps. Then we use steps in Sec.~\ref{sec:dmlmamm} to do MIL. Here we employ weights pretrained on the ImageNet due to the scarce of data. We use Adam optimization with learning rate $5 \times 10^{-5}$ for training models~\cite{ba2015adam}. The $\lambda$ for max pooling-based and label assignment-based MIL are $1 \times 10^{-5}$. The $\lambda$ and $\mu$ for sparse MIL are $5 \times 10^{-6}$ and $1 \times 10^{-5}$ respectively. For the label assignment-based MIL.

We firstly compare our methods to previous models validated on DDSM dataset and INbreast dataset in Table~\ref{tab:inbreast}. Previous hand-crafted feature-based methods require manually annotated detection bounding box or segmentation ground truth even in test denoting as manual~\cite{ball2007digital,varela2006use,domingues2012inbreast}. The feat. denotes requiring hand-crafted features. Pretrained CNN uses two CNNs to detect the mass region and segment the mass, followed by a third CNN to do  mass classification on the detected ROI region, which requires hand-crafted features to pretrain the network and needs multi-stages training\cite{dhungel2016automated}. Pretrained CNN+Random Forest further employs random forest and obtained 7\% improvement. These methods are either manually or need hand-crafted features or multi-stages training, while our methods are totally automated, do not require hand-crafted features or extra annotations even on training set, and can be trained in an end-to-end manner.
\begin{table}[t]
	\fontsize{9pt}{10pt}\selectfont\centering
	\caption{Accuracy Comparisons of the proposed deep MILs and related methods on test sets.}\label{tab:inbreast}
	\begin{tabular}{c|c|c|c|c}
		\hlinew{0.9pt}
		Methodology&Dataset&Set-up&Accu.&AUC\\		
		\hlinew{0.7pt}
		\tabincell{c}{Ball et al. \cite{ball2007digital}}&DDSM&Manual+feat.&0.87&N/A\\
		\hline \tabincell{c}{Varela et al. \cite{varela2006use}}&DDSM&Manual+feat.&0.81&N/A\\
		\hline \tabincell{c}{Domingues et al. \cite{domingues2012inbreast}}&INbr.&Manual+feat.&0.89&N/A\\
		\hline \tabincell{c}{Pretrained CNN \cite{dhungel2016automated}}&INbr.&Auto.+feat.&0.84$\pm{0.04}$&0.69$\pm{0.10}$\\
		\hline \tabincell{c}{Pretrained CNN+Random Forest \cite{dhungel2016automated}}&INbr.&Auto.+feat.&$\bf{0.91\pm{0.02}}$&0.76$\pm{0.23}$\\
		\hlinew{0.9pt}
		AlexNet &INbr.&Auto.&0.81$\pm{0.02}$&0.79$\pm{0.03}$\\
		\hline \tabincell{c}{AlexNet+Max Pooling MIL} &INbr.&Auto.&0.85$\pm{0.03}$&0.83$\pm{0.05}$\\
		\hline \tabincell{c}{AlexNet+Label Assign. MIL} &INbr.&Auto.&0.86$\pm{0.02}$&0.84$\pm{0.04}$\\
		\hline \tabincell{c}{AlexNet+Sparse MIL} &INbr.&Auto.&0.90$\pm{0.02}$&$\bf{0.89\pm{0.04}}$\\
		\hlinew{0.9pt}
	\end{tabular}
\end{table}

The max pooling-based deep MIL obtains better performance than the pretrained CNN using 3 different CNNs and detection/segmentation annotation in the training set. This shows the superiority of our end-to-end trained deep MIL for whole mammogram classification. According to the accuracy metric, the sparse deep MIL is better than the label assignment-based MIL, which is better than the max pooling-based MIL. This result is consistent with previous discussion that the sparsity assumption benefited from not having hard constraints of the label assignment assumption, which employs all the patches and is more efficient than max pooling assumption. Our sparse deep MIL achieves competitive accuracy to random forest-based pretrained CNN, while much higher AUC than previous work, which shows our method is more robust. The main reasons for the robust results using our models are as follows. Firstly, data augmentation is an important technique to increase scarce training datasets and proves useful here. Secondly, the transfer learning that employs the pretrained weights from ImageNet is effective for the INBreast dataset. Thirdly, our models fully explore all the patches to train our deep networks thereby eliminating any possibility of overlooking malignant patches by only considering a subset of patches. This is a distinct advantage over previous networks employing several stages.
\begin{figure}[!t]
	\setlength{\abovecaptionskip}{0.cm}
	\setlength{\belowcaptionskip}{-0.cm}
	\begin{center}
		\begin{minipage}{0.15\linewidth}
			\centerline{\includegraphics[width=2.5cm]{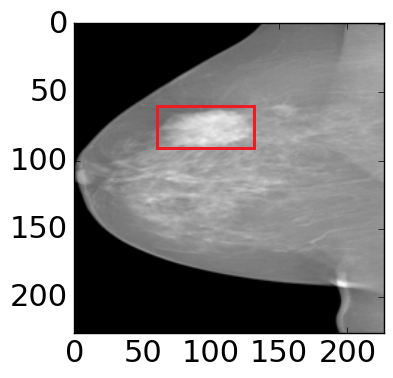}}
			\center{(a)}
		\end{minipage}
		\hspace{1cm}
		\begin{minipage}{0.15\linewidth}
			\centerline{\includegraphics[width=2.5cm]{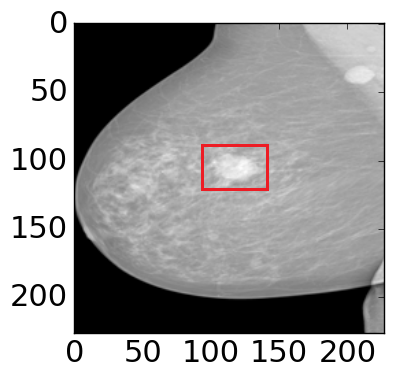}}
			\center{(b)}
		\end{minipage}
		\hspace{1cm}
		\begin{minipage}{0.15\linewidth}
			\centerline{\includegraphics[width=2.5cm]{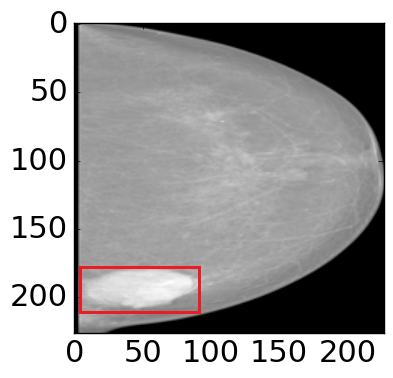}}
			\center{(c)}
		\end{minipage}
		\hspace{1cm}
		\begin{minipage}{0.15\linewidth}
			\centerline{\includegraphics[width=2.5cm]{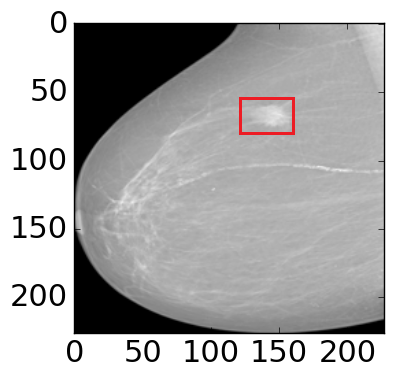}}
			\center{(d)}
		\end{minipage}
		\vfill
		\begin{minipage}{0.15\linewidth}
			\centerline{\includegraphics[width=2.5cm]{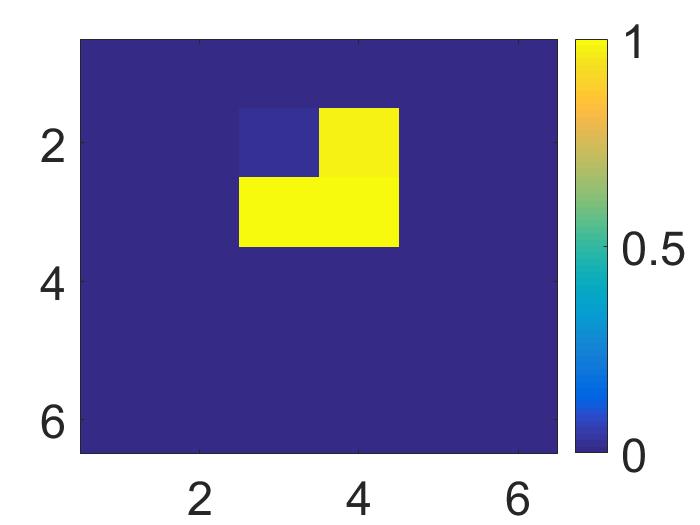}}
		\end{minipage}
		\hspace{1cm}
		\begin{minipage}{0.15\linewidth}
			\centerline{\includegraphics[width=2.5cm]{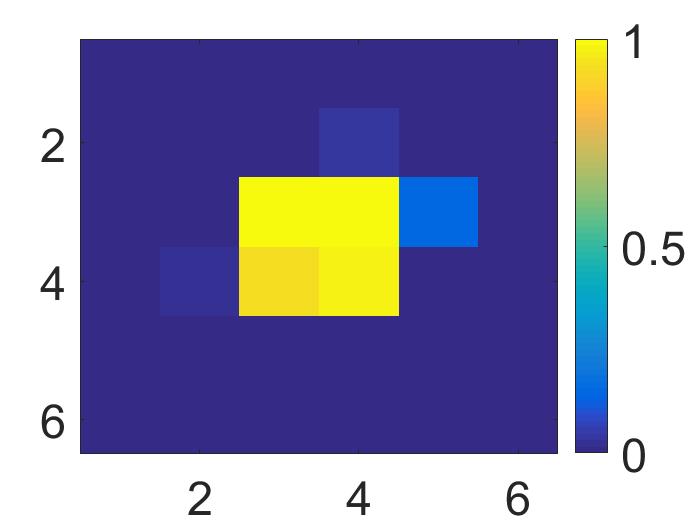}}
		\end{minipage}
		\hspace{1cm}
		\begin{minipage}{0.15\linewidth}
			\centerline{\includegraphics[width=2.5cm]{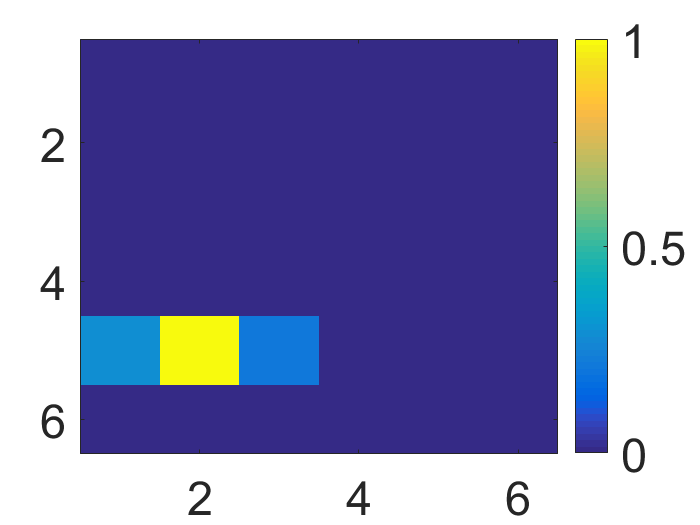}}
		\end{minipage}
		\hspace{1cm}
		\begin{minipage}{0.15\linewidth}
			\centerline{\includegraphics[width=2.5cm]{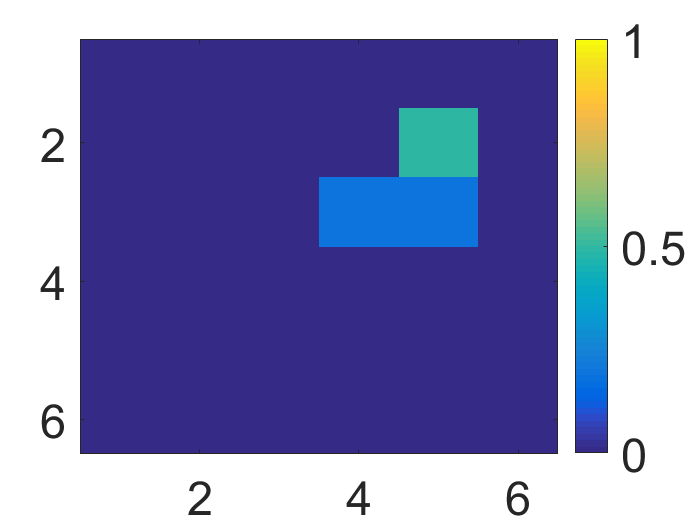}}
		\end{minipage}
		\vfill
		\begin{minipage}{0.15\linewidth}
			\centerline{\includegraphics[width=2.6cm]{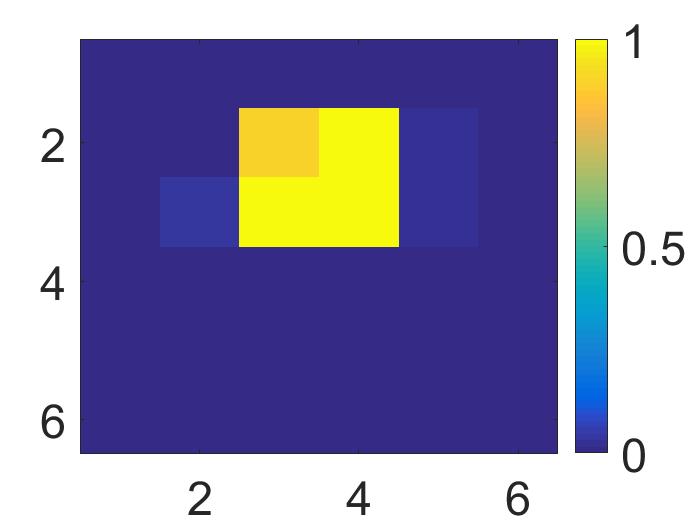}}
		\end{minipage}
		\hspace{1cm}
		\begin{minipage}{0.15\linewidth}
			\centerline{\includegraphics[width=2.6cm]{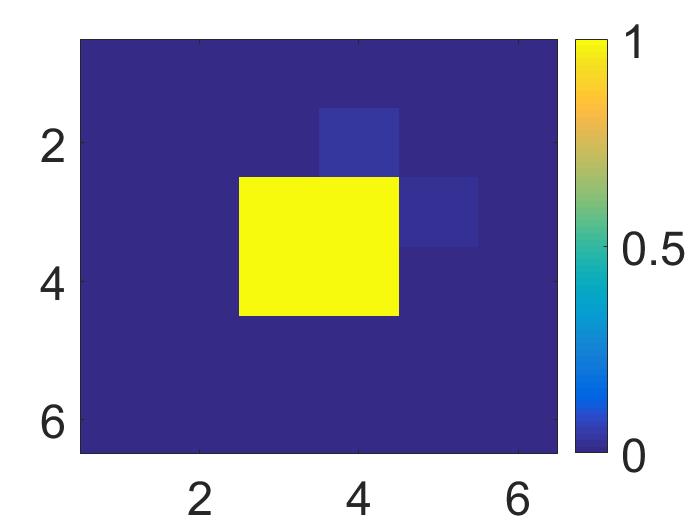}}
		\end{minipage}
		\hspace{1cm}
		\begin{minipage}{0.15\linewidth}
			\centerline{\includegraphics[width=2.6cm]{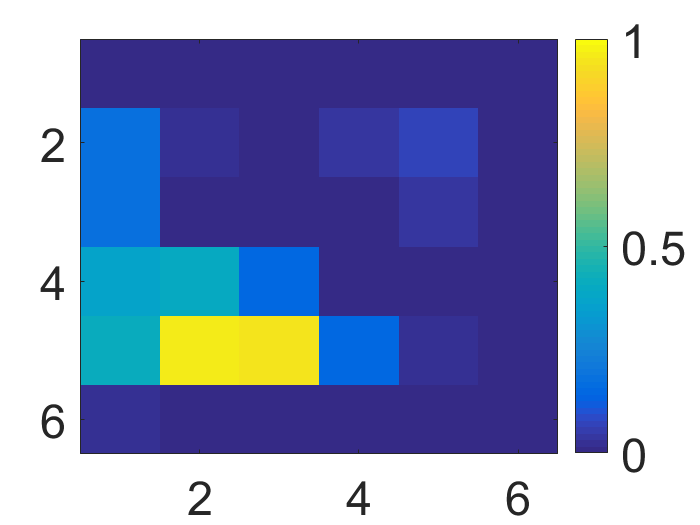}}
		\end{minipage}
		\hspace{1cm}
		\begin{minipage}{0.15\linewidth}
			\centerline{\includegraphics[width=2.6cm]{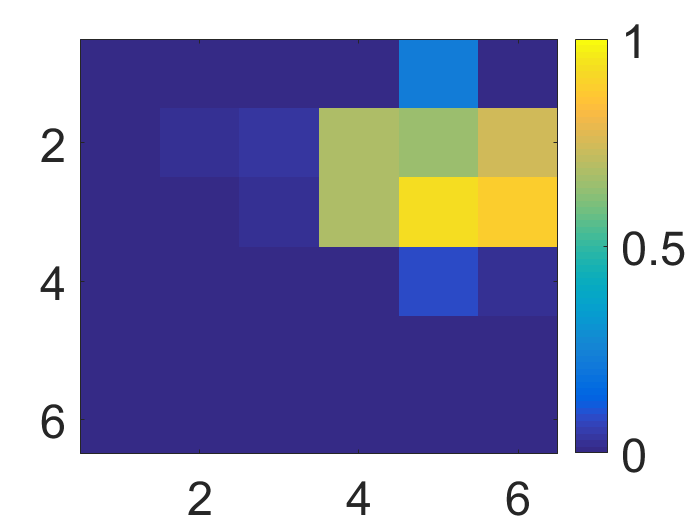}}
		\end{minipage}
		\vfill
		\begin{minipage}{0.15\linewidth}
			\centerline{\includegraphics[width=2.6cm]{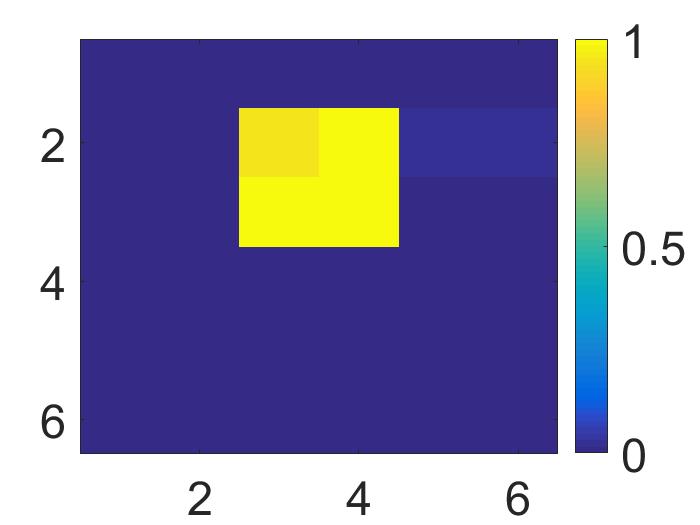}}
		\end{minipage}
		\hspace{1cm}
		\begin{minipage}{0.15\linewidth}
			\centerline{\includegraphics[width=2.6cm]{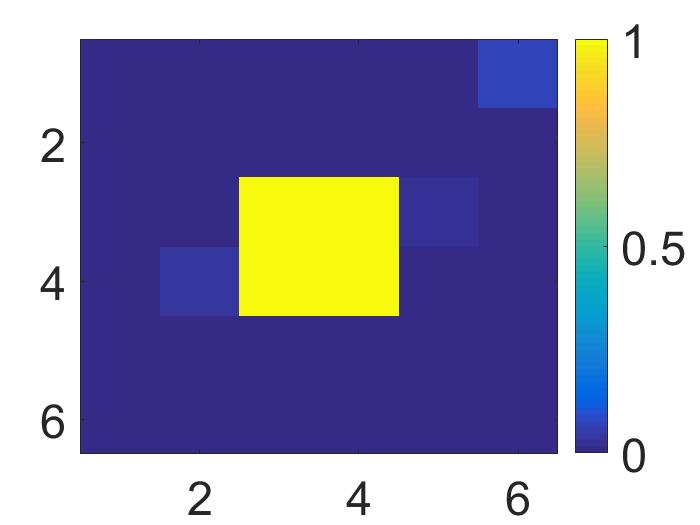}}
		\end{minipage}
		\hspace{1cm}
		\begin{minipage}{0.15\linewidth}
			\centerline{\includegraphics[width=2.6cm]{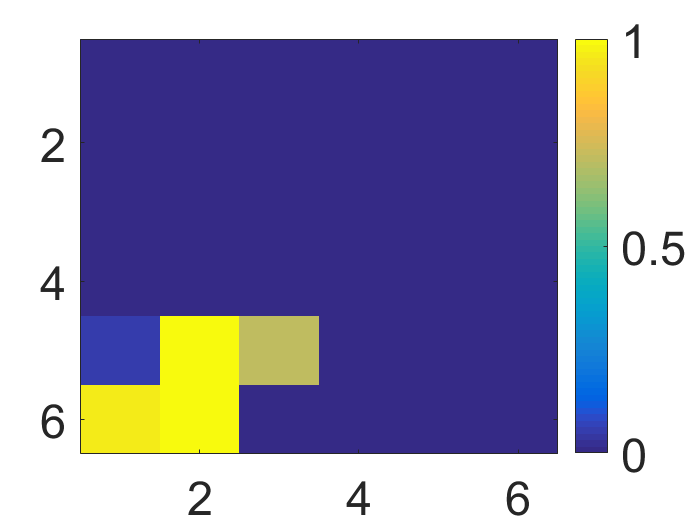}}
		\end{minipage}
		\hspace{1cm}
		\begin{minipage}{0.15\linewidth}
			\centerline{\includegraphics[width=2.6cm]{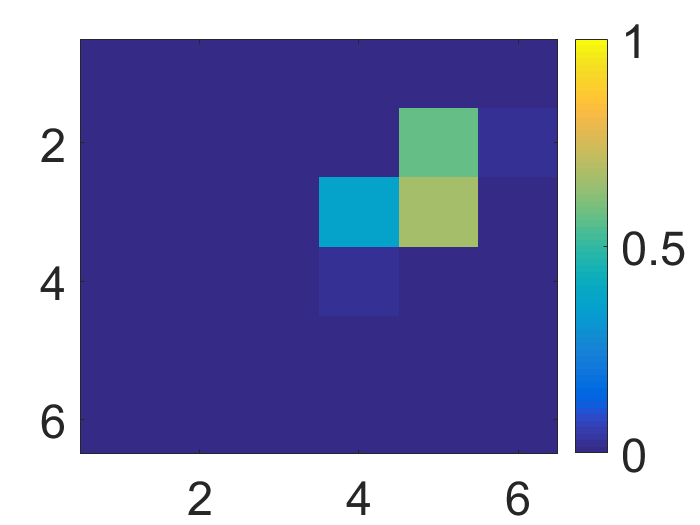}}
		\end{minipage}
		\caption{The visualization of predicted malignant probabilities for instances/patches in four resized mammograms. The first row is the resized mammogram. The red rectangle boxes are mass regions from the annotations on the dataset. The color images from the second row to the last row are the predicted malignant probability from logistic regression layer for (a) to (d) respectively, which are the malignant probabilities of patches/instances. Max pooling-based, label assignment-based, sparse deep MIL are in the second row, third row, fourth row respectively.}
		\label{fig:visresponse}
	\end{center}
\end{figure}

To further understand our deep MIL, we visualize the responses of logistic regression layer for four mammograms on test set, which represents the malignant probability of each patch, in Fig.~\ref{fig:visresponse}. We can see the deep MIL learns not only the prediction of whole mammogram, but also the prediction of malignant patches within the whole mammogram. Our models are able to learn the mass region of the whole mammogram without any explicit bounding box or segmentation ground truth annotation of training data. The max pooling-based deep multi-instance network misses some malignant patches in (a), (c) and (d). The possible reason is that it only considers the patch of max malignant probability in training and the model is not well learned for all patches. The label assignment-based deep MIL mis-classifies some patches in (d). The possible reason is that the model sets a constant $k$ for all the mammograms, which causes some mis-classifications for small masses. One of the potential applications of our work is that these deep MIL networks could be used to do weak mass annotation automatically, which provides evidence for the diagnosis.
\section{Conclusion}\label{sec:con}
In this paper, we propose end-to-end trained deep MIL for whole mammogram classification. Different from previous work using segmentation or detection annotations, we conduct mass classification based on whole mammogram directly. We convert the general MIL assumption to label assignment problem after ranking. Due to the sparsity of masses, sparse MIL is used for whole mammogram classification. Experimental results demonstrate more robust performance even without detection or segmentation annotation in the training. 

In future work, we plan to extend the current work by: 1) incorporating multi-scale modeling such as spatial pyramid to further improve whole mammogram classification, 2) employing the deep MIL to do annotation or provide potential malignant patches to assist diagnoses, and 3) applying to large datasets if the big dataset is available. 
\bibliographystyle{splncs03}
\small{\bibliography{typeinst}}

\begin{thebibliography}{10}
\bibitem{acs}
American cancer society. what are the key statistics about breast cancer?
\end{thebibliography}

\end{document}